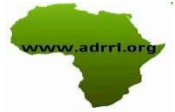

# Optimal Load Scheduling Using Genetic Algorithm to Improve the Load Profile

Farhat Iqbal[1], Shafiq ur Rehman[2] and Khawar Iqbal[3]
[1]Department of Physics, Sargodha University, Pakistan
[2]Departmet of Electrical Engineering, Sargodha University, Pakistan
[3]The University of Lahore, Sargodha, Pakistan
[1]**Correspondence**: farhat.iqbal@uos.edu.pk

## Abstract

Stability and protection of the electrical power systems are always of primary concern. Stability can be affected mostly by increase in the load demand. Power grids are overloaded in peak hours so more power generation units are required to cope the demand. Increase in power generation is not an optimal solution. With the enlargement in Smart grid (SG), it becomes easier to correlate the consumer demand and available power. The most significant featutre of smart grid is demand response (DR) which is used to match the demand of available electrical energy and shift the peak load into off peak hours to improve the economics of energy and stability of grid stations. Presently we used Genetic algorithm (GA) to schedule the load via real time pricing signal (RTP). Load is categorized depending on their energy requirement, operational constraint and duty cycle. We conclude that GA provides optimal solution for scheduling of house hold appliances by curtailing overall utilized energy cost and peak to average ratio hence improving the load profile.

# INTRODUCTION

Electrical power systems need to ensure the capability of matching demand and generation at distribution level. Residential load contributes towards the major load in overall energy consumption in almost all energy communication systems. Hussein S. & Boonruang M. (2018, February) concluded in their survey that 18% of entire energy is utilized by residential customer . Onur A. & BelginEmre T. (2017, December) described in their article that growing population of the consumer based needs demand to have new industries and markets hence escalating energy consumption day by day.According to Pradeepti L. & Mukesh K. (2015, August) investigations , when energy consumption demand is increased, it disturbs the grid stability in the form of cascaded blackout. It will be a uprising provocation task for power system designer to meet and regulate the increasing demand, while retaining the system stability. In that circumstance, actual consumption of available power and effective distribution of power to the consumer loads will be an significant task. A huge gap between power generation and demand, results higher cost of both generation and transmission. In addition, power generation unit are often not able to cope the load demands in peak hour (Dehgham S., Darafshian M., Shayanfar H. A. and Kazemi, 2011)resulting in additional disturbances, blackout, and instability . A grid should have a capability to accommodate instability errors caused by overloading(Deepak K. S., Rajiv S. and Prem K. K., 2010). An overloading in electrical system is an indication that some portion of the combined generation and transmission system has been operating beyond its capability mostly in peak hours. At the instant when power system itself in emergency state, it may shed partial load to sustain system integrity and traditional power grid do not have ability to tackle such problem (Chao R. C., Wen T. & Hua Y. C., 2011).

Many approaches have been developed to minimize the load curtailment without violating the system security limitations. Optimal load scheduling (OLS) is lively energetic energy management method for changing nature of user load. OLS is applied in such a way that it can cope the user demand as well as the grid stability by decreasing the peak to average ratio (PAR). OLS has lot of advantages i.e. gainful impact of cost reduction, reduced dependency of high octane fuels, higher costs and decreased carbon emission (Thillaina T. L., Dipti S. & Tang Z. S., 2012).Traditional grids are unable to respond fast due to their electro mechanical infrastructure, interruption and instability. Traditional grids are rapidly being transformed into smart grids (SG) for renewable energy integration of local generation plants and demand side management (Usman M. Khalid & Nadeem J., 2018) and (Pedram S., Hamed M., Vincent W. W. & Robert S., 2013).

SG offers two-way power flow i.e. from generation units via distribution system to consumer and vice versa. Smart meters are utilized to prevent resulting cascaded blackout.Smart meters receive pricing signal from utility company and sent to energy management controller(EMC). Energy management controller schedules the appliances according to pricing signal (Tamal R., Avijit D. & Zhen N. , 2007).

Demand response (DR) is a key feature of OLS scheme in smart grid (SG). DR is the strategy, which help utility company and user to shed unnecessary load during peak hours in a day. The EMC installed at the user end introduces economic incentives to the consumer, hence motivates the consumer to switch their appliances running schedule form highly peak hours to off peak hours (Pavithra N. & Priya B., 2017). The EMC takes decision based on pricing signal. There are different pricing schemes used by utility company for load scheduling such as, time-of-use pricing scheme (TOH), day-ahead pricing scheme (DAH), flat pricing scheme (FP), real time pricing scheme (RTP) and critical peak pricing scheme ( Jacqueline C., Katherine W., Chialin C., 2018). All of the schemes have their own advantages while RTP reflect the true picture of low cost energy availability in a system. The RTP horizon distributed in numerous slots during a day, off peak, mid peak and on peak. Furthermore, mid peak and on peak is dependent on seasons. Neelam B. & Sandeep K.(2018) used a mixed integer linear programming (MILP) for shifting of home appliances from peak hours to off peak houres to curtail the PAR and consumption cost.

Nikhil D. R. and Nand k. K.(2017) applied A Particle Swarm Optimization approach for the scheduling of home appliance via demand side management (DSM). In this research 4-distribution transformer were used to feed electrical power to 200 consumers Although 14 type of shiftable appliances are used while effective utilization of energy management purely depends on these shiftable appliances. The limitation of this technique is its large computational complexity and convergence time.

Previously a geographical information system has been applied to display the data from power grid to geographical map for 7600 houses by Monika, Dipti S. & Thomas R.(2017). The basic pillars of geographical mapping system are presenting business tier and data tier. The purpose of real-time imaging is to display pattern of energy consumption and energy availability to regulate the demand curve. The limitation of such technique is that it is suitable only for large number of consumers. Presently we applied Genetic algorithm (GA) to schedule the load via RTP. Load is categorized depending on their energy requirement, operational constraint and duty cycle. GA provides optimal solution for scheduling of house hold appliances by curtailing overall energy consumption cost and PAR and improves the load profile.

**System Model**

We catagoriesed residential loads into three types as 1. necessary load (NL i.eTelevision, Laptop, Computer, Tube lights, Bulbs and other Illumination etc.), 2.consistent load (CL i.e. air conditioner & Refrigerator, water pumps, oven, iron) and 3. inconsistent load (ICL i.e. Washing machine, and Microwave oven).The nature of CL and ICL is controllable and NL is independent. Residential load can be scheduled based on various time slots and operational constraints. The load are scheduled to decrease the power consumption in peak hours and increase power avaiablity in off peak hours.

The minimum required load NL is neither shift able nor under the control of energy management controller with DR program. CL runs in off peak hours with minimum cost specially when local photo voltaic unit (LPVU) is available, CL must cope their duty cycle whenever these are turned ON. ICL are directly associated with DR program. We consider a self-governing home equipped with local PV unit. Rahim S.-et-al (2015) formulated knapsack method using similar approach to minimize the electricity bill. States of appliances with respect to load classification used in present study are shown in figure 1. In first stage, appliances are classified according to consumption of power and we do not take supply from the LPVU. The, NL can be swtiched ON in any horizon of time of the day, while CL would be turn ON according to their duty cycle. Although once CL will turn ON, it will not beturned OFF till it completes its duty cycle. The appliances that are considered ICL have interruptible duty cycle. ICL can be turned OFF and ON in their duty cycle. In second stage, when consumer turns ON any appliance, firstly controller will check that if LPVU is capable to energize load then energy shell be provided from LPVU. Such an optimization problem has been solved partially by genetic algorithm with LPVU and without LPVU(Kun W., Huining L., Sabita M., Yan Z. and Song G., 2018).On the other hand, if LPVU is not capable to cope the energy demand then power would be taken from grid. Our core objective is to transfer the peak load from grid to LPVU when price signal is maximum and to curtail the overall electricity bill. Proposed OLS model is presented in figure: 2.

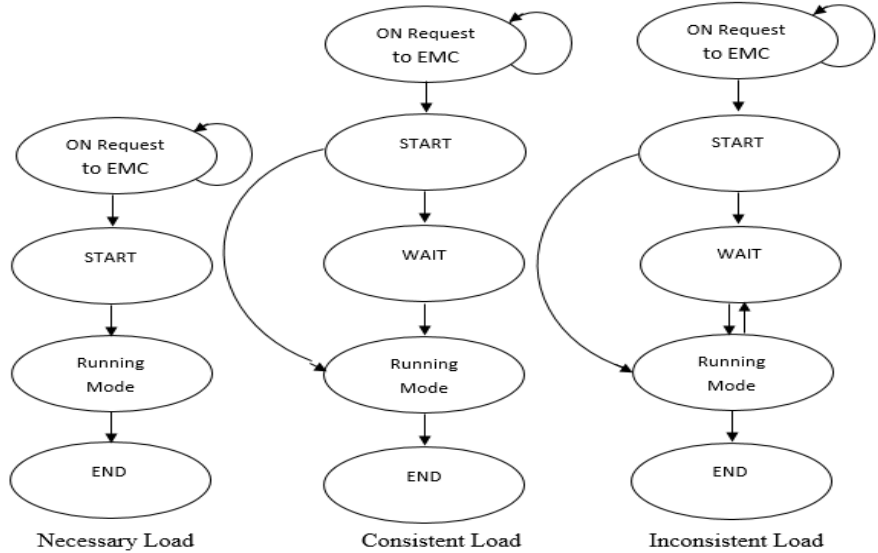

**Figure 1. Turn ON/OFF States of Appliances**

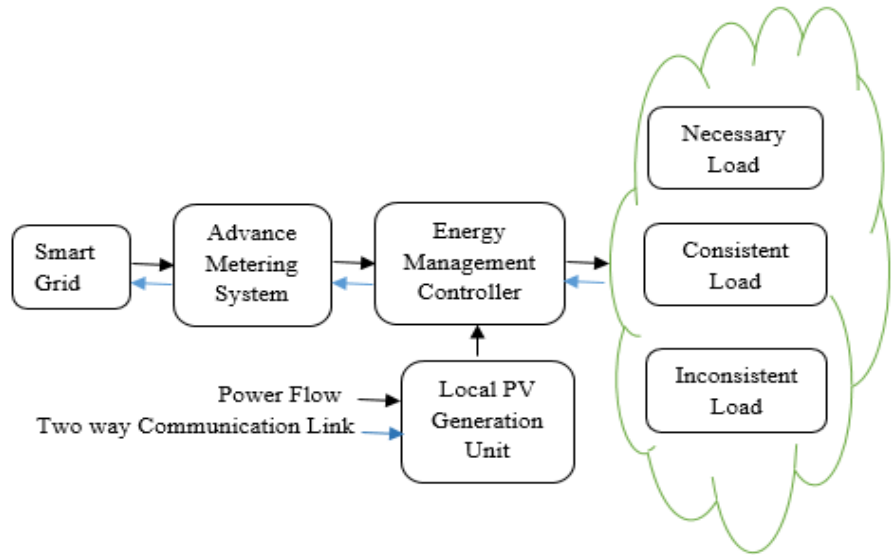

**Figure 2. OLS Model**

**Model of overall Utilization of Energy:**
Energy management controller (EMC) takes decision, and schedules the appliance's turn ON time and turn OFF time based on price signal. The cost is calculated on the basis of turn ON period and total energy consumption as given in the equation 1,

$$E^T = \sum_{t=0}^{24} \sum_a^y \sum_b^z R_{ab}(t) \times E_{ab}(t) \quad (1)$$

where,
$E_{ab}$  Utilized energy of appliance a of type b
t       Time horizon of a day
a       Total number of appliances
b       Type of appliance
$R_{ab}$  ON/OFF state of appliance a of type b

**Model of Energy Utilization Cost:**
We consider that cost of price signal for respective time is real in nature. The overall cost of utilized energy is estimated by multiplication of real time price signal ρ(t) with utilized energy by consumer in time span. Although overall billing cost with respect to utilized energy by consumer is given as,

$$C^T = \sum_{t=0}^{24} \sum_a^y \sum_b^z R_{ab}(t) \times E_{ab}(t) \times \rho(t) \quad (2)$$

Where, $C^T$ is energy utilization cost and ρ(t) price signal of utilized time slot.

**Model of local PV unit:**

Considering that LPVU source is available. Consumer feeds by mix-energy via grid and PV panel, which lead to overcome the overall cost of consumed energy. Energy produced by PV cells depend upon sunlight, while efficiency of PV cell is more in summer and less in winters. Power produced by PV panel is dependant on solar radiation, area of panel covered by sun light and direction of radiation. The energy produced by PV panel is expressed as,

$$\in(t) = 10 \times \frac{1}{\sqrt{2\pi}\sigma} e^{\left(\frac{(t-\delta)^2}{2\sigma^2}\right)} \qquad (3)$$

Where,
$$\begin{cases} \in(t) > 0 \\ \in(t) < 0 \end{cases} \begin{array}{l} \text{at day} \\ \text{at night} \end{array}$$

$\in(t)$   Generation of solar energy.
$\sigma$   Variance
$\delta$   Mean distribution
$t$   Time horizon of a day

**PAR Calculation:**

Normally more demand is observed at grid throughout peak hours and less demand of energy observed in off peak hours. It is necessary to decline the power demand during peak hours, its is in the best interest of the utility companies and consumer also. During peak hour, the magnitude of price signal will be more and during off peak hour's magnitude of price signal will be less. The average to peak ratio estimated by peak load and average load of the day, which is given in (4).

$$\upsilon = \left\{ \frac{\max(R(t))}{\frac{1}{24}\sum_0^{24} R(t)} \right\} \qquad (4)$$

**Objective Function:**

In order to reduce the PAR and billing price, load should be schedules in different time slots. We used Knap technique (Rahim S.-et-al , 2015)to formulate the scheduling of load at different times. Hence  forecasting the boundary about demand vs utilization of energy for consumers to reduce the PAR. The problem of energy utilization cost model is expressed as below Reduce to

$$\sum_{t=0}^{24} \sum_i^m \sum_j^n R_{ij}(t) \times E_{ij}(t) \times \rho(t) \qquad (5)$$

Subject to

$$\sum_{t=0}^{24} \sum_i^m \sum_j^n R_{ij}(t) \times E_{ij}(t) \leq D(t) \qquad (6)$$

$$\sum_{t=0}^{24} \sum_i^m \sum_j^n R_{ij}(t) = OC_{ij} \qquad (7)$$

Where $D(t)$ is the maximum demand limit in specific time horizon, while $OC_{cd}$ is total number of ON calls of respective appliance c type d.

$$\beta_i \leq 24 - OC_i \qquad \forall\ CL \qquad (8)$$
$$t_i^s \leq \beta_i \leq t_i^s + OC_i \qquad \forall\ ICL \qquad (9)$$
$$\beta_i = t_i^s \forall\ NL \qquad (10)$$
$$\upsilon_s < \upsilon_{us} \qquad (11)$$
$$E_S^T = E_{uS}^T \qquad (12)$$
$$E(t) - \in(t) > 0 \qquad (13)$$
$$R \in (0,1)$$

Where,
$\upsilon_s$   Peak schedule load
$\upsilon_{us}$   Peak un-schedule Load
$E_S^T$   Scheduled utilized energy
$E_{uS}^T$   Un-scheduled utilized energy
$\beta_i$   Operational horizon of appliance

**Proposed Algorithm for OLS**
Genetic algorithm globally known as search based method, which has the capability to schedule the load to reach an optimal solution. In genetic algorithm, our major concern is about chromosomes, because population of chromosomes provide optimal solution regarding the suggested problem. Random population is generated in the form of binary array, which represent the total quantity of appliances. Each bit represent the status of each appliance. We have total number of N smart appliance and M chromosomes (M=N).

**Initialization of population**: The size of initial population depends on consumer priorities of appliance usage. As the population size increases, accuracy and complexity of algorithm also increases. The initial population is characterized by MxN.

**Fitness function:** In this stage, fitness score is estimated for individual chromosome as we have bit array [1 0 1 0 1 0 1], that show the ON/FF state of appliance. In given array there are seven bits, state of appliance number 1, 3, 5, 7 are ON and state of Appliance number 2, 4, 6 are OFF. Total bill estimated based on state of appliances and summation of energy consumption of each appliance. Fitness function is calculate as follow as:

$$\sum_{t=0}^{24} \sum_{a}^{y} \sum_{b}^{z} R_{ab}(t) \times E_{ab}(t) \times \rho(t) \quad (14)$$

**Selection:** Probability of the most fit chromosomes and fitness score is evaluated based on previous generation. Selection is performed one time in the algorithm from incumbent generation based on fitness score of chromosomes and offspring developed. There are many techniques, which have been used for selection of chromosomes i.e. Roulette Wheel Selection, Rank Selection, Boltzmann Selection, and Tournament Selection. In this work, we used tournament selection method it has high flexibility and selection pressure can be changed easily by tuning the tournament size.

**Crossover:** Crossover is actually a genetic operator, which uses to combine the feature of two parents in new generated chromosome. The new chromosomes isgenerated through the process of crossover and its survival depends on parent's fitness value. In this work, we used one point crossover method as. schedulingof household appliances is discrete in nature .Increasing the cross ratewill result in highier convergence rate.

**Mutation:** The mutation operator sustains the genetic diversity in the initial population and mutation effects convergence. If mutation degree is amplified, the convergence will slow. In this work, we used binary mutation method.

**Termination:** Searching algorithm determines the condition when it will end and cycle repeat until the best fitness value achieved. The termination condition depends on preset value of objective function, number of generations, the process terminates when no further improvement found in the previous generation.

**RESULTS AND DISCUSSIONS**

In this section, outcomes are discussed in term of power consumption, billing cost and peak to average ratio. Simulations outcomes of a self-governing house consist of different types of load The different type of load are discussed in table 1. NL's are not scheduled; they are supposed to be in running condition throughout the day. However, ICL and CL are scheduled according to cost signal. CL must remain ON until the process is completed and gets ON only once in 24 hour cycle, while the duty cycle of ICL can be disturbed any time during operation. The cost signal is assumed to be a real time pricing (RTP) signal.

Table 1: Electrical load ratting & Specification

| Load Type | Number of Appliances | Loads | Operational Hours | Ratting (KWH) |
|---|---|---|---|---|
| NL | 2 | Laod-1 | 22 | 1.5 |
|  |  | Laod-2 | 23 | 0.5 |
| CL | 2 | Laod-3 | 5 | 6 |
|  |  | Laod-4 | 5 | 1.5 |
| ICL | 2 | Laod-5 | 7 | 3.5 |
|  |  | Laod-6 | 8 | 1.4 |

The cost defined for each hour in a day and shown in Figure 3. which illustrates that during time interval 11 to 14 energy price signal is highier. Genetic algorithm schedule the load in time slots 1 to 10 and 15 to 24 to decrease the consumption cost. The reduction of PAR with the integration of LPVU, means the load classification and incorporation of LPVU are used into action simultaneously. This research work is carried out by the use of solar energy however it is adaptable to any form of energy and its generation calculation can is obtained from equation 3.2.

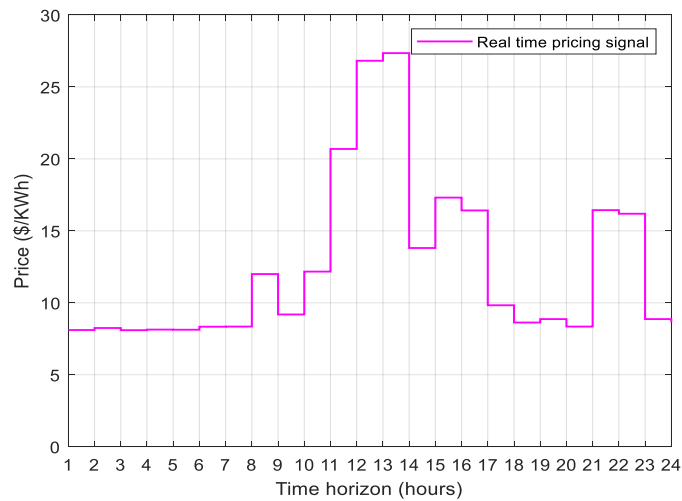

Figure 3: Real Time Pricing Signal

Figure 4 shows that the generated power form LPVU in respective horizon, while simulations indicate that generation of power based on temperature and irradiations. Moreover, at night generation by solar power is considered to be zero. Load is scheduled based on real time pricing signal classification of load, while figure 5 illustrates the consumed energy profile of self-governing home with and without incorporation of renewable energy source. The EMC manages the load according to RTP signal and classification of load, which ultimately leads to the economical energy sources. The EMC transfer the load from peak hours to off peak hours, which leads to the reduction of energy consumption cost and effective utilization of energy sources.

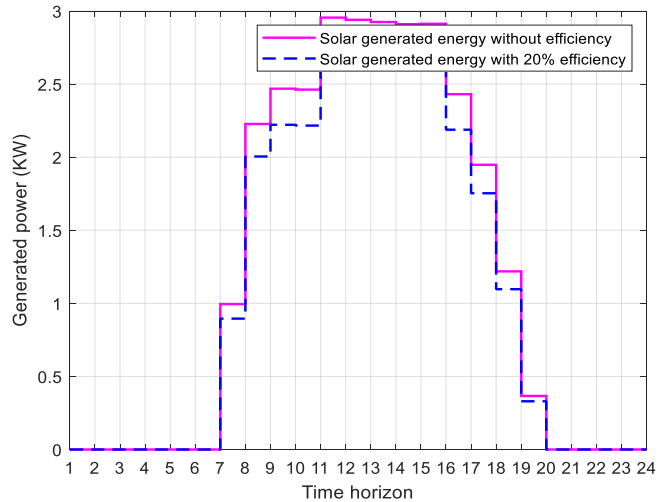

Figure 4: LPVU generation in a day.

Although, the total utilized energy of scheduled and un-scheduled load does not differ. As shown in figure that with the integration of local photo voltaic unit EMC controller shares that energy with grid.

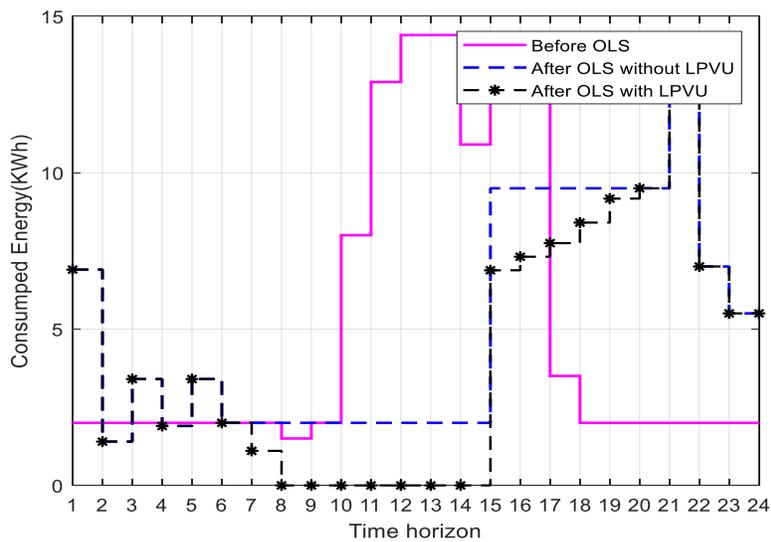

Figure 5: Consumed energy profile of a day.

The daily electricity consumption cost profile shown in figure 6, which illustrate the significantreduction in bills in case of scheduling without LPVU availability and scheduling with LPVU. Therefore, these outcomes show that the recommended algorithm performs well in all circumstances. Therefore, this scheme refers that load cataloging and incorporation of LPVU is helpful for both user and utility and has huge scope in an EMC program.Result shows that billing cost reduced to 7.10 % without incorporation of LPVU, while billing cost is reduced to 55.62% with incorporation of LPVU. Figure 7 illustrates a clear PAR decline which implies a major impact on grid stability and consistency.

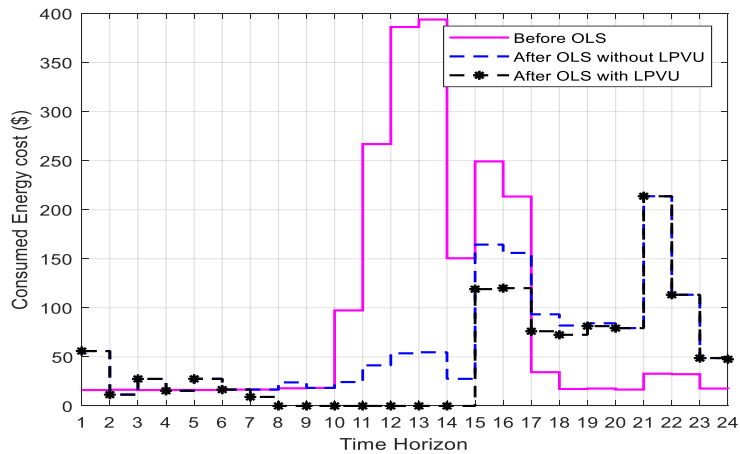

Figure 6: Consumed energy cost of a day.

The PAR reduction is directly proportional to reduction of cost value for upcoming time space, which implies that with decline in PAR optimal economics of system is achieve.

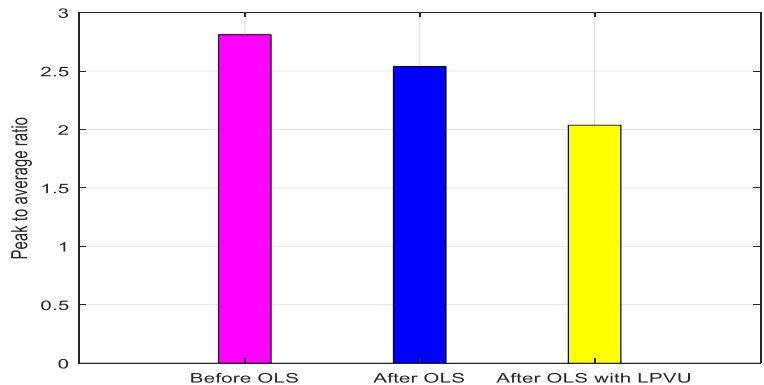

Figure 7: PAR profile.

Table 2: Bill and Cost profile

| | |
|---|---|
| Total Bill reduction without LPVU (%) | 7.1068 |
| Total Bill reduction with LPVU (%) | 55.62 |
| PAR Reduction without LPVU (%) | 9.72 |
| PAR Reduction with LPVU (%) | 34.08 |

Table 2 illustrate clearly the PAR decline and energy consumption cost. More prominent PAR have reduced due to the availability of LPVU and billing cost is reduced more due to availability of LPVU.

**CONCLUSION AND FUTURE WORK**

The research aims to curtail the energy utilization cost and PAR, which is in the best interest of both utility and consumer. The algorithm adopted in the research for LPVU availability, and it can fedthe load to its full capacity. If LPVU cannot sustain the load, then the EMC schedules the load considering the RTP. Results obtained through simulations suggest that cost of energy and PAR value reduced to great extent. Results demonstrated that cost of electricity after adaption GA is reduced. In future, research work can be extended to several stations on commercial level.

**REFERENCES**

Swalehe, H., & Marungsri, B. (2018). Intelligent Algorithm for Optimal Load Management in Smart Home Appliance Scheduling in Distribution System. 2018 International Electrical Engineering Congress (iEECON), 1–4. Krabi, Thailand: IEEE. https://doi.org/10.1109/ieecon.2018.8712166

Ayan , O., & Emre Turkay, B. (2017, December). Smart home energy management technologies based on demand side management and a review of smart plugs with smart thermostats, 1247–1252. Presented at the 2017 10th International Conference on Electrical and Electronics Engineering (ELECO), Bursa, Turkey. Retrieved from https://ieeexplore.ieee.org/document/8266290

Lakra, P., & Kirar, M. (2015). Load Shedding Techniques for System with Co-generation: A Review. Electrical and Electronics Engineering: An International Journal (ELELIJ) , 4(6), 1–14. https://doi.org/10.14810/elelij.2015.4307

Dehghan, S., Darafshian-Maram, M., Shayanfar, H., & Kazemi, A. (2011). Optimal load shedding to enhance voltage stability and voltage profile based on a multi-objective optimization technique. 2011 IEEE Trondheim PowerTech, 1–5. Trondheim, Norway: IEEE. https://doi.org/10.1109/ptc.2011.6019310

Singh, D., Shekhar, R., & Kalra, P. (2010). Optimal load shedding: An economic approach. TENCON 2010 - 2010 IEEE Region 10 Conference, 1–4. Fukuoka, Japan: IEEE. https://doi.org/10.1109/tencon.2010.5686763

Chen, C., Wen-Ta Tsai, Chen, H., Ching-Ying Lee, Chun-Ju Chen, & Hong-Wei Lan. (2011). Optimal load shedding planning with genetic algorithm. 2011 IEEE Industry Applications Society Annual Meeting, 1–6. Orlando, USA: IEEE. https://doi.org/10.1109/ias.2011.6074299

Logenthiran, T., Srinivasan, D., & Shun, T. Z. (2012). Demand Side Management in Smart Grid Using Heuristic Optimization. IEEE Transactions on Smart Grid, 3(3), 1244–1252. https://doi.org/10.1109/tsg.2012.2195686

Khalid, M., & Javaid, N. (2018). An optimal scheduling of smart home appliances using heuristic techniques with real-time coordination. 2018 1st International Conference on Power, Energy and Smart Grid (ICPESG), 1–6. Mirpur Azad Kashmir, Pakistan: IEEE. https://doi.org/10.1109/icpesg.2018.8384505

Samadi, P., Mohsenian-Rad, H., Wong, V. W. S., & Schober, R. (2013). Tackling the Load Uncertainty Challenges for Energy Consumption Scheduling in Smart Grid. IEEE Transactions on Smart Grid, 4(2), 1007–1016. https://doi.org/10.1109/tsg.2012.2234769

Roy, T., Das, A., & Ni, Z. (2017). Optimization in load scheduling of a residential community using dynamic pricing. 2017 IEEE Power & Energy Society Innovative Smart Grid Technologies Conference (ISGT), 1–5. Washington, DC, USA: IEEE. https://doi.org/10.1109/isgt.2017.8086087

Pavithra, N., & Esther, B. (2017). Residential demand response using genetic algorithm. 2017 Innovations in Power and Advanced Computing Technologies (i-PACT), 1–4. Vellore, India: IEEE. https://doi.org/10.1109/ipact.2017.8245143

Corbett, J., Wardle, K., & Chen, C. (2018). Toward a sustainable modern electricity grid: The effects of smart metering and program investments on demand-side management performance in the US electricity sector 2009-2012. IEEE

Transactions on Engineering Management, 65(2), 252–263. https://doi.org/10.1109/tem.2017.2785315

Bhati, N., & Kakran, S. (2018). Optimal Household Appliances Scheduling Considering Time Based Pricing Scheme. 2018 International Conference on Power Energy, Environment and Intelligent Control (PEEIC), 717–721. Greater Noida, India: IEEE. https://doi.org/10.1109/peeic.2018.8665487

Rahate, N., & Kinhekar, N. (2017). Demand side management for household equipment's. 2017 International Conference on Information, Communication, Instrumentation and Control (ICICIC), 1–5. Indore, India: IEEE. https://doi.org/10.1109/icomicon.2017.8279108

Monika, Srinivasan, D., & Reindl, T. (2017). Demand side management in residential areas using geographical information system. 2017 IEEE Conference on Energy Internet and Energy System Integration (EI2), 1–6. Beijing, China: IEEE. https://doi.org/10.1109/ei2.2017.8245766

Rahim, S., Khan, S., Javaid, N., Shaheen, N., Iqbal, Z., & Rehman, G. (2015). Towards Multiple Knapsack Problem Approach for Home Energy Management in Smart Grid. 2015 18th International Conference on Network-Based Information Systems, 1–5. Taipei, Taiwan: IEEE. https://doi.org/10.1109/nbis.2015.11

Wang, K., Li, H., Maharjan, S., Zhang, Y., & Guo, S. (2018). Green Energy Scheduling for Demand Side Management in the Smart Grid. IEEE Transactions on Green Communications and Networking, 2(2), 596–611. https://doi.org/10.1109/tgcn.2018.2797533